\documentclass[runningheads]{llncs}
\usepackage{subcaption}
\usepackage{graphicx}
\usepackage{amsmath,amssymb} 
\usepackage{commath}
\usepackage{color}
\usepackage{cite}
\usepackage{epstopdf}
\usepackage{float}
\usepackage{array}
\usepackage{enumerate}
\usepackage{hyperref}
\usepackage[width=122mm,left=12mm,paperwidth=146mm,height=193mm,top=12mm,paperheight=217mm]{geometry}
\begin{document}
\pagestyle{headings}
\mainmatter

\title{Human pose estimation via Convolutional Part Heatmap Regression} 

\titlerunning{Human pose estimation via Convolutional Part Heatmap Regression}

\authorrunning{Adrian Bulat \and Georgios Tzimiropoulos}

\author{
	Adrian Bulat 
	\and 
    Georgios Tzimiropoulos
}


\institute{Computer Vision Laboratory, University of Nottingham
\email{\{adrian.bulat,yorgos.tzimiropoulos\}@nottingham.ac.uk}
}

\maketitle

\begin{abstract}
This paper is on human pose estimation using Convolutional Neural Networks. Our main contribution is a CNN cascaded architecture specifically designed for learning part relationships and spatial context, and robustly inferring pose even for the case of severe part occlusions. To this end, we propose a detection-followed-by-regression CNN cascade. The first part of our cascade outputs part detection heatmaps and the second part performs regression on these heatmaps. The benefits of the proposed architecture are multi-fold: It guides the network where to focus in the image and effectively encodes part constraints and context. More importantly, it can effectively cope with occlusions because part detection heatmaps for occluded parts provide low confidence scores which subsequently guide the regression part of our network to rely on contextual information in order to predict the location of these parts. Additionally, we show that the proposed cascade is flexible enough to readily allow the integration of various CNN architectures for both detection and regression, including recent ones based on residual learning. Finally, we illustrate that our cascade achieves top performance on the MPII and LSP data sets. Code can be downloaded from  \url{http://www.cs.nott.ac.uk/~psxab5/}

\end{abstract}

\begin{figure}
\centering 
\includegraphics[scale=0.35]{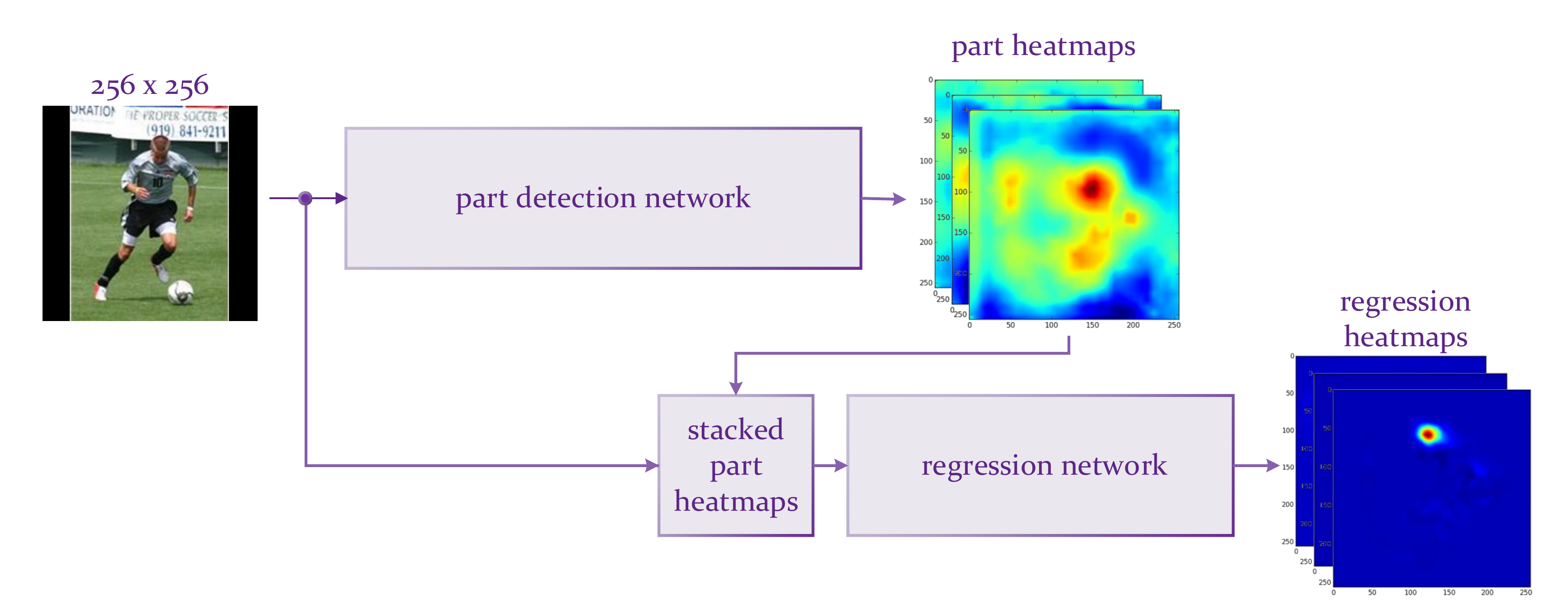}
\caption{\textbf{Proposed architecture}: Our CNN cascade consists of two connected deep subnetworks. The first one (upper part in the figure) is a part detection network trained to detect the individual body parts using a per-pixel sigmoid loss. Its output is a set of $N$ part heatmaps. The second one is a regression subnetwork that jointly regresses the part  heatmaps stacked along with the input image to confidence maps representing the location of the body parts.}
\label{fig:OurNetworkFArchFull}
\end{figure}  

\begin{figure*}
    \centering
	\subcaptionbox*{occluded ankle}{\includegraphics[height=1in]{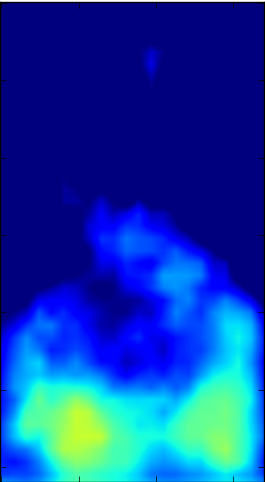}}
	\subcaptionbox*{occluded knee}{\includegraphics[height=1in]{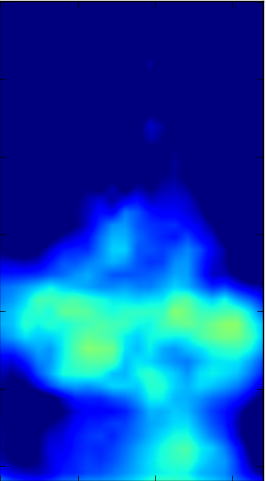}}
	\subcaptionbox*{visible knee}{\includegraphics[height=1in]{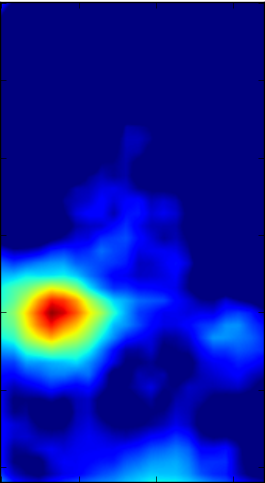}}
	\subcaptionbox*{occluded wrist}{\includegraphics[height=1in]{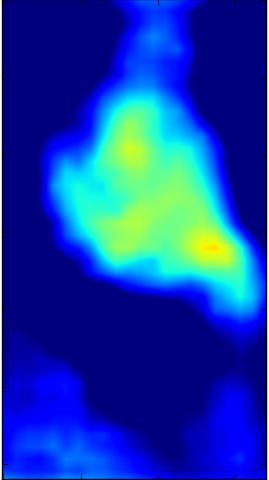}}
	\subcaptionbox*{neck}{\includegraphics[height=1in]{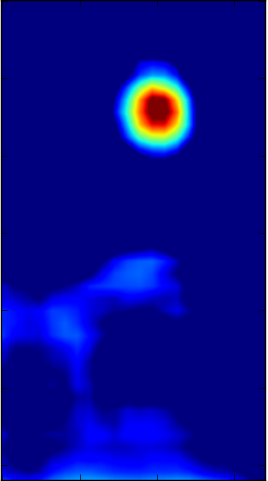}}
	\subcaptionbox*{head}{\includegraphics[height=1in]{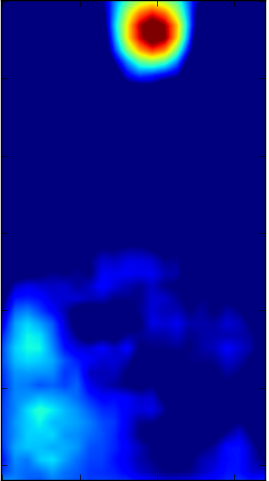}}
	\subcaptionbox*{result}{\includegraphics[height=1in]{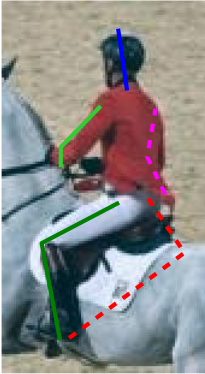}}
	
	\centering
	\subcaptionbox*{}{\includegraphics[height=1in]{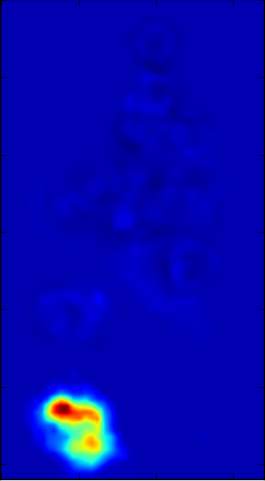}}
	\subcaptionbox*{}{\includegraphics[height=1in]{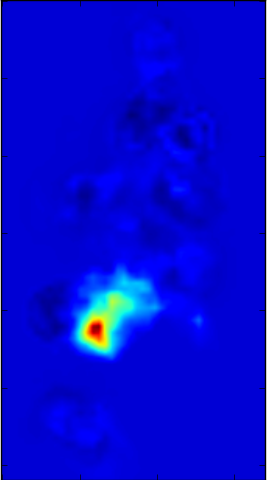}}
	\subcaptionbox*{}{\includegraphics[height=1in]{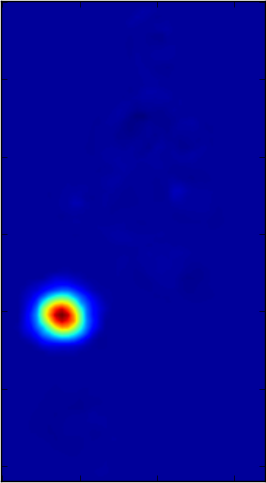}}
	\subcaptionbox*{}{\includegraphics[height=1in]{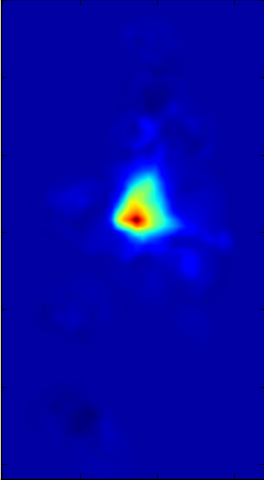}}
	\subcaptionbox*{}{\includegraphics[height=1in]{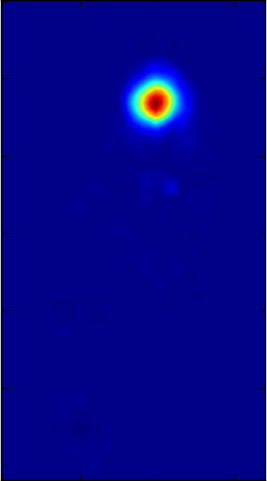}}
	\subcaptionbox*{}{\includegraphics[height=1in]{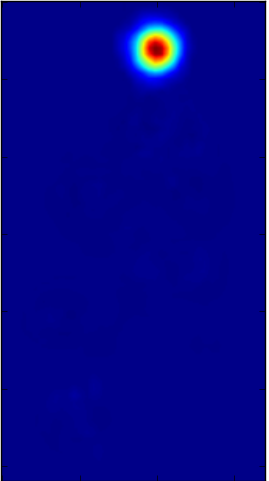}}
	\subcaptionbox*{}{\includegraphics[height=1in]{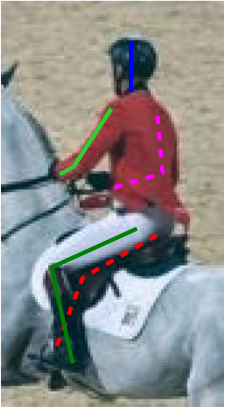}}	
	
	\caption{\textbf{Paper's main idea}: The first row shows the produced part detection heatmaps for both visible (neck, head, left knee) and occluded (ankle, wrist, right knee) parts (drawn with a dashed line). Observe that the confidence for the occluded parts is much lower than that of the non-occluded parts but still higher than that of the background providing useful context about their rough location. The second row shows the output of our regression subnetwork. Observe that the confidence for the visible parts is higher and more localized and clearly the network is able to provide high confidence for the correct location of the occluded parts. \textbf{Note}: image taken from LSP test set.}
\label{fig:idea}
\end{figure*}

\section{Introduction}
Articulated human pose estimation from images is a Computer Vision problem of extraordinary difficulty. Algorithms have to deal with the very large number of feasible human poses, large changes in human appearance (e.g. foreshortening, clothing), part occlusions (including self-occlusions) and the presence of multiple people within close proximity to each other. A key question for addressing these problems is how to extract strong low and mid-level appearance features capturing discriminative as well as relevant contextual information and how to model complex part relationships allowing for effective yet efficient pose inference. Being capable of performing these tasks in an end-to-end fashion, Convolutional Neural Networks (CNNs) have been recently shown to feature remarkably robust performance and high part localization accuracy. Yet, the accurate estimation of the locations of occluded body parts is still considered a difficult open problem. The main contribution of this paper is a CNN cascaded architecture specifically designed to alleviate this problem.  

There is a very large amount of work on the problem of human pose estimation. Prior to the advent of neural networks most prior work was primarily based on pictorial structures \cite{felzenszwalb2005pictorial} which model the human body as a collection of rigid templates and a set of pairwise potentials taking the form of a tree structure, thus allowing for efficient and exact inference at test time. Recent work includes sophisticated extensions like mixture, hierarchical, multimodal and strong appearance models \cite{yang2011articulated, pishchulin2013poselet, tian2012exploring, sapp2013modec, pishchulin2013strong}, non-tree models \cite{karlinsky2012using, dantone2013human} as well as cascaded/sequential prediction models like pose machines \cite{ramakrishna2014pose}.

More recently methods based on Convolutional Neural Networks have been shown to produce remarkable performance for a variety of difficult Computer Vision tasks including recognition \cite{he2016deep, simonyan2014very}, detection \cite{girshick2015fast} and semantic segmentation \cite{long2015fully} outperforming prior work by a large margin. A key feature of these approaches is that they integrate non-linear hierarchical feature extraction with the classification or regression task in hand being also able to capitalize on very large data sets that are now readily available. In the context of human pose estimation, it is natural to formulate the problem as a regression one in which CNN features are regressed in order to provide joint prediction of the body parts \cite{toshev2014deeppose, pfister2014deep, pfister2015flowing, belagiannis2015robust}. For the case of non-visible parts though, learning the complex mapping from occluded part appearances to part locations is hard and the network has to rely on contextual information (provided from the other visible parts) to infer the occluded parts' location. In this paper, we show how to circumvent this problem by proposing a detection-followed-by-regression CNN cascade for articulated human pose estimation. 

\subsection{Main Contribution}

The proposed architecture is a CNN cascade consisting of two  components (see Fig. \ref{fig:OurNetworkFArchFull}): the first component (part detection network) is a deep network for part detection that produces detection heatmaps, one for each part of the human body. We train part detectors jointly using  pixelwise sigmoid cross entropy loss function \cite{zhang2015fine}. The second component is a deep regression subnetwork that jointly regresses the location of all parts (both visible and occluded), trained via confidence map regression \cite{pfister2015flowing}. Besides the two subnetworks, the key feature of the proposed architecture is the input to the regression subnetwork: we propose to use a stacked representation comprising the part heatmaps produced by the detection network. The proposed representation guides the network where to focus and encodes structural part relationships. Additionally, our cascade does not suffer from the problem of regressing occluded part appearances: because the part heatmaps for the occluded parts provide low confidence scores, they subsequently guide the regression part of our network to rely on contextual information (provided by the remaining parts) in order to predict the location of these parts. See Fig. \ref{fig:idea} for a graphical representation of our paper's main idea. The proposed cascade is very simple, can be trained end-to-end, and is flexible enough to readily allow the integration of various CNN architectures for both our detection and regression subnetworks. To this end, we illustrate two instances of our cascade, one based on the more traditional VGG converted to fully convolutional (FCN) \cite{simonyan2014very, long2015fully} and one based on residual learning \cite{he2016deep, newell2016stacked}. Both architectures achieve top performance on both MPII \cite{andriluka20142d} and LSP \cite{johnson2010clustered} data sets.

\section{Closely Related Work}
\label{S:Related}

\textbf{Overview of prior work.} Recently proposed methods for articulated human pose estimation using CNNs can be classified as detection-based \cite{chen2014articulated, tompson2014joint, tompson2015efficient, pishchulin2015deepcut, insafutdinov2016deepercut} or regression-based \cite{toshev2014deeppose, pfister2014deep, pfister2015flowing, belagiannis2015robust, carreira2015human, wei2016convolutional}. Detection-based methods are relying on powerful CNN-based part detectors which are then combined using a graphical model \cite{chen2014articulated, tompson2014joint} or refined using regression \cite{tompson2015efficient, pishchulin2015deepcut}. Regression-based methods try to learn a mapping from image and CNN features to part locations. A notable development has been the replacement of the standard L2 loss between the predicted and ground truth part locations with the so-called confidence map regression which defines an L2 loss between predicted and ground truth confidence maps encoded as 2D Gaussians centered at the part locations \cite{tompson2014joint, pfister2015flowing} (these regression confidence maps are not to be confused with the part detection heatmaps proposed in our work). As a mapping from CNN features to part locations might be difficult to learn in one shot, regression-based methods can be also applied sequentially (i.e. in a cascaded manner) \cite{toshev2014deeppose, carreira2015human, wei2016convolutional}. Our CNN cascade is based on a two-step detection-followed-by-regression approach (see Fig. \ref{fig:OurNetworkFArchFull}) and as such is related to both detection-based \cite{tompson2015efficient, pishchulin2015deepcut} and regression-based methods \cite{pfister2015flowing, carreira2015human, wei2016convolutional}.  

\textbf{Relation to regression-based methods.} Our detection-followed-by-regression cascade is related to \cite{pfister2015flowing} which can be seen as a two-step regression-followed-by-regression approach. As a first step \cite{pfister2015flowing} performs confidence map regression (based on an L2 loss) as opposed to our part detection step which is learnt via pixelwise sigmoid cross entropy loss. Then, in \cite{pfister2015flowing} pre-confidence maps are used as input to a subsequent regression network. We empirically found that such maps are too localised providing small spatial support. In contrast, our part heatmaps can provide large spatial context for regression. For comparison purposes, we implemented the idea of \cite{pfister2015flowing} using two different architectures, one based on VGG-FCN and one on residual learning, and show that the proposed detection-followed-by-regression cascade outperforms it for both cases (see section \ref{S:analysis}). In order to improve performance, regression methods applied in a sequential, cascaded fashion have been recently proposed in \cite{carreira2015human, wei2016convolutional}. In particular, \cite{wei2016convolutional} has recently reported outstanding results on both LSP \cite{johnson2010clustered} and MPII \cite{andriluka20142d} data sets using a six-stage CNN cascade.

\textbf{Relation to detection-based methods.} Regarding detection-based methods, \cite{pishchulin2015deepcut} has produced state-of-the-art results on both MPII and LSP data sets using a VGG-FCN network \cite{simonyan2014very, long2015fully} to detect the body parts along with an L2 loss for regression that refines the part prediction. Hence, \cite{pishchulin2015deepcut} does not include a subsequent part heatmap regression network as our method does. The work of \cite{tompson2015efficient} uses a part detection network as a first step in order to provide crude estimates for the part locations. Subsequently, CNN features are cropped around these estimates and used for refinement using regression. Hence, \cite{tompson2015efficient} does not include a subsequent part heatmap regression network as our method does, and hence does not account for contextual information but allows only for local refinement. 

\textbf{Residual learning.} Notably, all the aforementioned methods were developed prior to the advent of residual learning \cite{he2016deep}. Very recently, residual learning was applied for the problem of human pose estimation in \cite{insafutdinov2016deepercut} and \cite{newell2016stacked}. Residual learning was used for part detection in the system of \cite{insafutdinov2016deepercut}. The ``stacked hourglass network'' of \cite{newell2016stacked} elegantly extends FCN \cite{long2015fully} and deconvolution nets \cite{zeiler2011adaptive} within residual learning, also allowing for a more sophisticated and heavy processing during top-down processing. We explore residual learning within the proposed CNN cascade; notably for our residual regression subnetwork, we used a single ``hourglass network'' \cite{newell2016stacked}.

\section{Method}

The proposed part heatmap regression is a CNN cascade illustrated in Fig. \ref{fig:OurNetworkFArchFull}. Our cascade consists of two connected subnetworks. The first subnetwork is a part detection network trained to detect the individual body parts using a per-pixel softmax loss. The output of this network is a set of $N$ part detection heatmaps. The second subnetwork is a regression subnetwork that jointly regresses the part detection heatmaps stacked with the image/CNN features to confidence maps representing the location of the body parts. 

We implemented two instances of part heatmap regression: in the first one, both subnetworks are based on VGG-FCN \cite{simonyan2014very, long2015fully} and in the second one, on residual learning \cite{he2016deep, newell2016stacked}. For both cases, the subnetworks and their training are described in detail in the following subsections. The following paragraphs outline important details about the training of the cascade, and are actually independent of the architecture used (VGG-FCN or residual). 

\textbf{Part detection subnetwork.} While \cite{long2015fully} uses a per-pixel softmax loss encoding different classes with different numeric levels, in practice, for the human body this is suboptimal because the parts are usually within close proximity  to each other, having high chance of overlapping. Therefore, we follow an approach similar to \cite{zhang2015fine} and encode part label information as a set of $N$ binary maps, one for each part, in which the values within a certain radius around the provided ground truth location are set to 1 and the values for the remaining background are set to 0. This way, we thus tackle the problem of having multiple parts in the very same region. Note that the detection network is trained using visible parts only, which is fundamentally different from the previous regression-based approaches\cite{pfister2015flowing,tompson2014joint,tompson2015efficient}.

The radius defining ``correct location'' was selected so that the targeted body part is fully included inside. Empirically, we determined that a value of 10px to be optimal for a body size of 200px of an upright standing person.

We train our body part detectors jointly using pixelwise sigmoid cross entropy loss function:

\begin{equation}
	 l_1 = \dfrac{1}{N} \sum\limits_{n=1}^{N} \sum\limits_{i=1}^{W} \sum\limits_{j=1}^{H}[p_{ij}^n\log \hat{p_{ij}^n}+(1-p_{ij}^n)\log(1-\hat{p_{ij}^n})],
\end{equation}
where $ p_{ij}^n $ denotes the ground truth map of the $ n $th part at pixel location $ (i,j) $ (constructed as described above) and $ \hat{p_{ij}^n} $ is the corresponding sigmoid output at the same location.

\textbf{Regression subnetwork.} While the detectors alone provide good performance, they lack a strong relationship model that is required to improve (a) accuracy and (b) robustness particularly required in situations where specific parts are occluded. To this end, we propose an additional subnetwork that jointly regresses the location of all parts (both visible and occluded). The input of this subnetwork is a multi-channel representation produced by stacking the $N$ heatmaps produced by the part detection subnetwork, along with the input image. (see Fig. \ref{fig:OurNetworkFArchFull}). This multichannel representation guides the network where to focus and encodes structural part relationships. Additionally, it ensures that our network does not suffer from the problem of regressing occluded part appearances: because the part detection heatmaps for the occluded parts provide low confidence scores, they subsequently guide the regression part of our network to rely on contextual information (provided by the remaining parts) in order to predict the location of these parts. 

The goal of our regression subnetwork is to predict the points' location via regression. However, direct regression of the points is a difficult and highly non-linear problem caused mainly by the fact that only one single correct value needs to be predicted. We address this by following a simpler alternative route \cite{tompson2014joint, pfister2015flowing}, regressing a set of confidence maps located in the immediate vicinity of the correct location (instead of regressing a single value). The ground truth consists of a set of $ N $ layers, one for each part, in which the correct location of each part, be it visible or not is represented by Gaussian with a standard deviation of 5px.

We train our subnetwork to regress the location of all parts jointly using the following L2 loss:
 
\begin{equation}
	l_2 = \dfrac{1}{N} \sum\limits_{n=1}^{N} \sum\limits_{ij} \norm{\widetilde{M}_{n}(i,j)-M_n(i,j)}^2, 
\end{equation}
where $ \widetilde{M}_{n}(i,j)$ and $ M_n (i,j)$ represent the predicted and the ground truth confidence maps at pixel location $ (i,j) $ for the $ n $th part, respectively.

\subsection{VGG-FCN part heatmap regression}

\begin{figure}
\centering 
\includegraphics[scale=0.23]{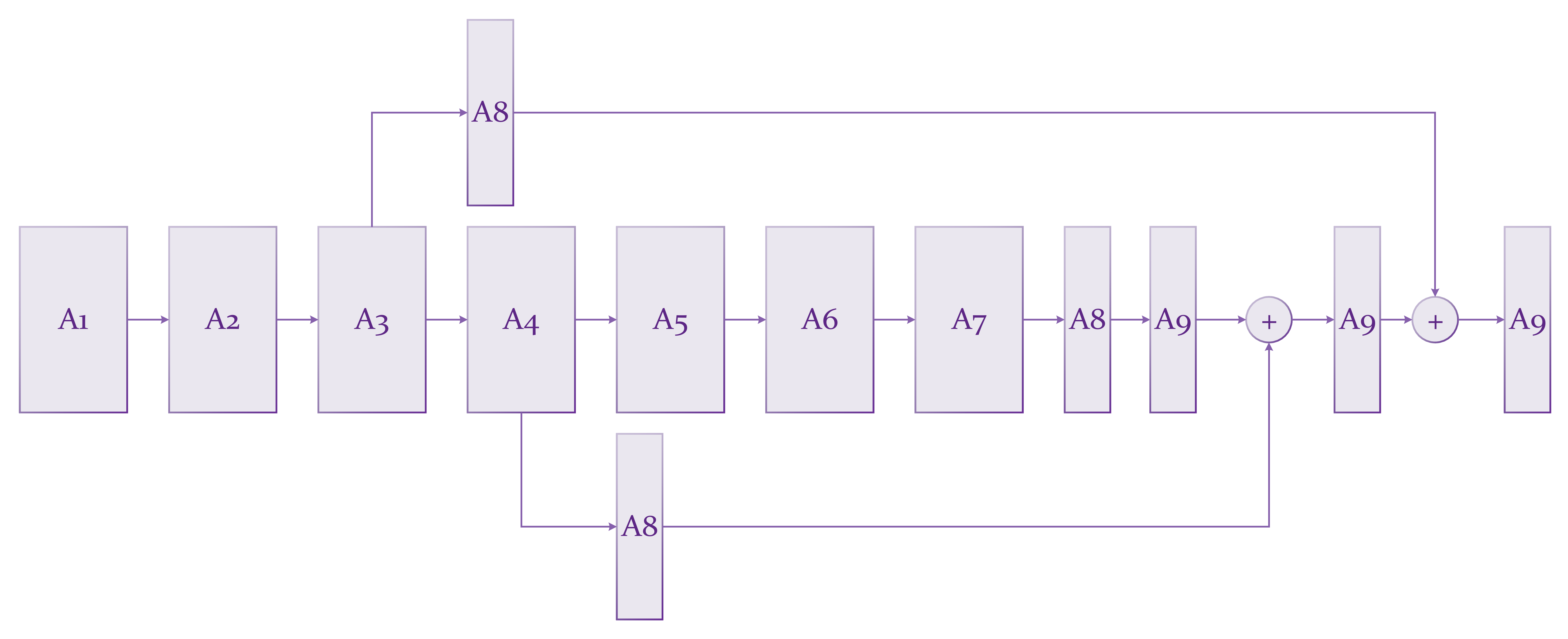}
\caption{The VGG-FCN subnetwork used for body part detection. The blocks A1-A9 are defined in Table \ref{table:DetectionContentVGG}.}
\label{fig:OurDetNetworkVGG}
\end{figure}

\textbf{Part detection subnetwork.} We based our part detection network architecture on the VGG-16 network \cite{simonyan2014very} converted to fully convolutional by replacing the fully connected layers with convolutional layers of kernel size of 1 \cite{long2015fully}. Because the localization accuracy offered by the 32px stride is insufficient, we make use of the entire algorithm as in \cite{long2015fully} by combining the earlier level CNN features, thus reducing the stride to 8px. For convenience, the network is shown in Fig. \ref{fig:OurDetNetworkVGG} and Table \ref{table:DetectionContentVGG}.

\begin{table}
	\begin{center}
		\scriptsize
		\caption{Block specification for the VGG-FCN part detection subnetwork. Torch notations (channels, kernel, stride) and (kernel, stride) are used to define the conv and pooling layers.}
		\label{table:DetectionContentVGG}
		\begin{tabular}{| *9{>{\raggedright\arraybackslash}p{1.3cm}|}}
        \hline
			A1 & A2 &A3 & A4 & A5 & A6  & A7 & A8 & A9 \\ \hline
		 2x conv layer (64, 3x3, 1x1), pooling & 2x conv layer (128, 3x3, 1x1), pooling  & 3x conv layer (256, 3x3, 1x1), pooling  & 3x conv layer (512, 3x3, 1x1), pooling & 3X conv layer(512, 1x1, 1x1), pooling  &  conv layer (4096, 7x7, 1x1) & conv layer (4096, 1x1, 1x1)& conv layer(16, 1x1, 1x1) & bilinear upsample \\ \hline
		\end{tabular}
	\end{center}
\end{table}
\setlength{\tabcolsep}{1.4pt}

\begin{figure}
\centering 
\includegraphics[scale=0.27]{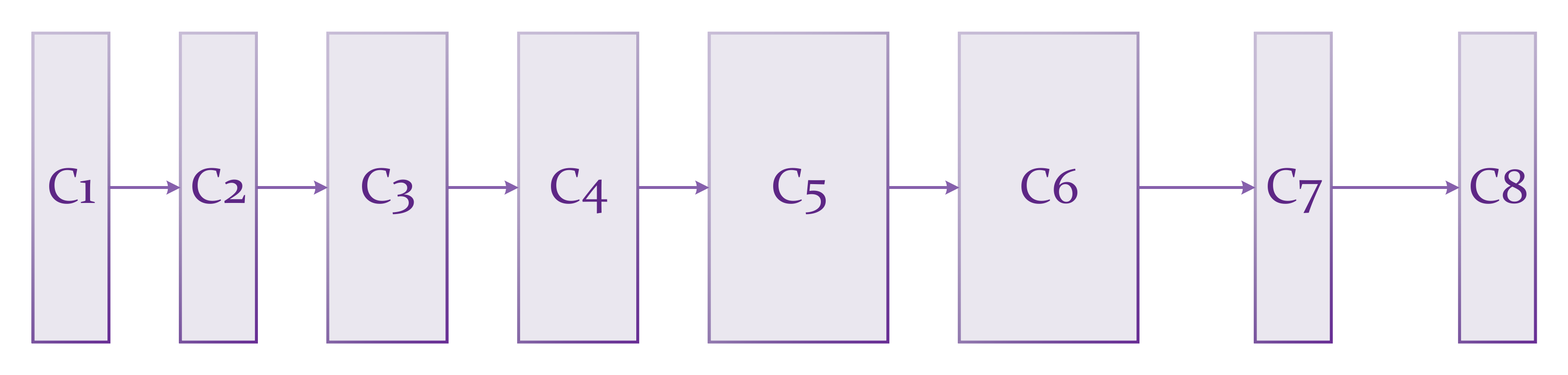}
\caption{The VGG-based subnetwork used for regression. The blocks C1-C8 are defined in Table \ref{table:RegressionContentVGG}.}
\label{fig:OurRegNetworkVGG}
\end{figure}

\textbf{Regression subnetwork.} We have chosen a very simple architecture for our regression sub-network, consisting of 7 convolutional layers. The network is shown in Fig. \ref{fig:OurRegNetworkVGG} and Table \ref{table:RegressionContentVGG}. The first 4 of these layers use a large kernel size that varies from 7 to 15, in order to capture a sufficient local context and to increase the receptive field size which is crucial for learning long-term relationships. The last 3 layers have a kernel size equal to 1.

\textbf{Training.} For training on MPII, all images were cropped after centering on the person and then scaled such that a standing-up human has height 300px. All images were resized to a resolution of 380x380px. To avoid overfitting, we performed image flipping, scaling (between 0.7 and 1.3) and rotation (between -40 and 40 degrees). Both rotation and scaling were applied using a set of predefined step sizes. Training the network is a straightforward process. We started by first training the body part detection network, fine-tuning from VGG-16 \cite{simonyan2014very, long2015fully} pre-trained on ImageNet \cite{deng2009imagenet}. The detectors were then trained for about 20 epochs using a learning rate progressively decreasing from $ 1e-8 $ to $ 1e-9 $.  For the regression subnetwork, all layers were initialized with a Gaussian distribution (std=0.01). To accelerate the training and avoid early divergence we froze the training of the detector layers, training only the subnetwork. We let this train for 20 epochs with a learning rate of $ 0.00001 $ and then $ 0.000001 $. We then trained jointly both networks for 10 epochs. We  found that one can train both the part detection network and the regression subnetwork jointly, right from the beginning, however, the aforementioned approach results in faster training. 

For LSP, we fine-tuned the network for 10 epochs on the 1000 images of the training set. Because LSP provides the ground truth for only 14 key points, during fine-tuning we experimented with two different strategies: (i) generating the points artificially and (ii) stopping the backpropagation for the missing points. The later approach produced better results overall. The training was done using the caffe\cite{jia2014caffe} bindings for Torch7\cite{collobert2011torch7}.

\begin{table}
	\begin{center}
		\caption{Block specification for the VGG-based regression subnetwork. Torch notations (channels, kernel, stride) and (kernel, stride) are used to define the conv and pooling layers.}
        \scriptsize
		\label{table:RegressionContentVGG}
		\begin{tabular}{| *8{>{\raggedright\arraybackslash}p{1.5cm}|}}
        \hline
			C1 & C2 & C3 & C4 & C5 & C6  & C7 & C8  \\ \hline
		 conv layer(64, 9x9, 1x1) & conv layer(64, 13x13, 1x1)  & conv layer(128, 13x13, 1x1)  & conv layer(256, 15x15, 1x1) & conv layer(512, 1x1, 1x1)  &  conv layer(512, 1x1, 1x1) & conv layer(16, 1x1, 1x1) & deconv layer (16, 8x8, 4x4) \\ \hline
		\end{tabular}
	\end{center}
\end{table}
\setlength{\tabcolsep}{1.4pt}

\subsection{Residual part heatmap regression} \label{sec:det}

\begin{figure}
\centering 
\includegraphics[scale=0.4]{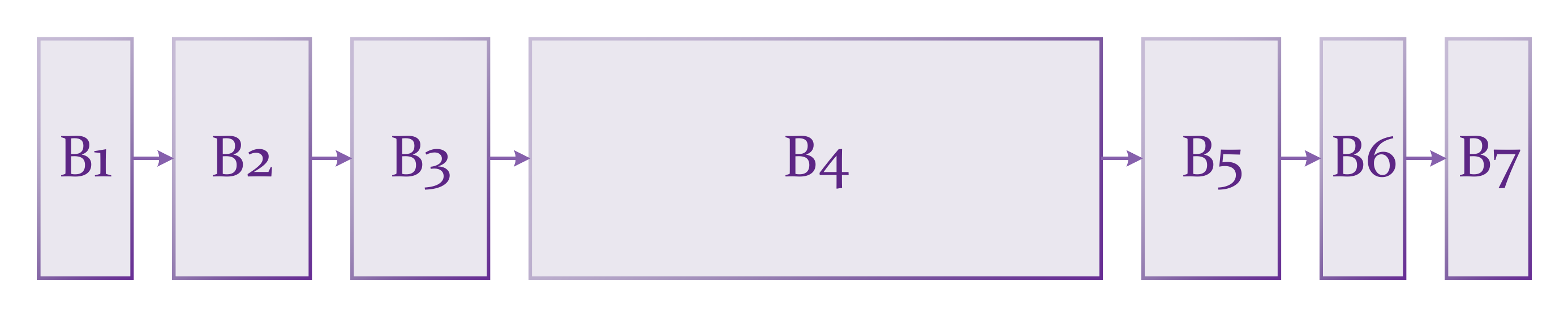}
\caption{The architecture of the residual part detection subnetwork. The network is based on ResNet-152 and its composing blocks. The blocks B1-B7 are defined in Table \ref{table:DetContent}. See also text.}
\label{fig:OurDetNetwork}
\end{figure}

\textbf{Part detection subnetwork.} Motivated by recent developments in image recognition \cite{he2016deep}, we used ResNet-152 as a base network for part detection. Doing so requires making the network able to make predictions at pixel level which is a relative straightforward process (similar ways to do this are described in \cite{insafutdinov2016deepercut, wu2016high, dai2016r}). The network is shown in Fig. \ref{fig:OurDetNetwork} and Table \ref{table:DetContent}. Blocks B1-B4 are the same as the ones in the original ResNet, and B5 was slightly modified. We firstly removed both the fully connected layer after B5 and then the preceding average pooling layer. Then, we added a scoring convolutional layer B6 with $N$ outputs, one for each part. Next, to address the extremely low output resolution, we firstly modified B5 by changing the stride of its convolutional layers from 2px to 1px and then added (after B6) a deconvolution \cite{zeiler2011adaptive} layer B7 with a kernel size and stride of 4, that upsamples the output layers to match the resolution of the input. We argue that for our detection subnetwork, knowing the exact part location is not needed. All added layers were initialized with 0 and trained using rmsprop \cite{tieleman2012lecture}.

\begin{table}
	\begin{center}
		\caption{Block specification for the residual part detection network. Torch notations (channels, kernel, stride) and (kernel, stride) are used to define the conv and pooling layers. The bottleneck modules are defined as in \cite{he2016deep}.}
        \scriptsize
		\label{table:DetContent}
		\begin{tabular}{| *7{>{\raggedright\arraybackslash}p{1.7cm}|}}
        \hline
			B1 & B2 & B3 & B4 & B5 & B6  & B7  \\ \hline
		 1x conv layer (64,7x7,2x2) 1x pooling (3x3, 2x2)   & 3x bottleneck modules [(64,1x1), (64,3x3), (256,1x1)]   & 8x bottleneck modules [(128,1x1), (128,3x3), (512,1x1)]  & 38x bottleneck modules [(256,1x1), (256,3x3), (1024,1x1)] & 3x bottleneck modules [(512,1x1), (512,3x3), (2048,1x1)] &  1x conv layer (16,1x1,1x1) & 1x deconv layer (16,4x4,4x4)  \\ \hline
		\end{tabular}
	\end{center}
\end{table}
\setlength{\tabcolsep}{1.4pt}

\begin{figure}[t]
\centering 
\includegraphics[scale=0.25]{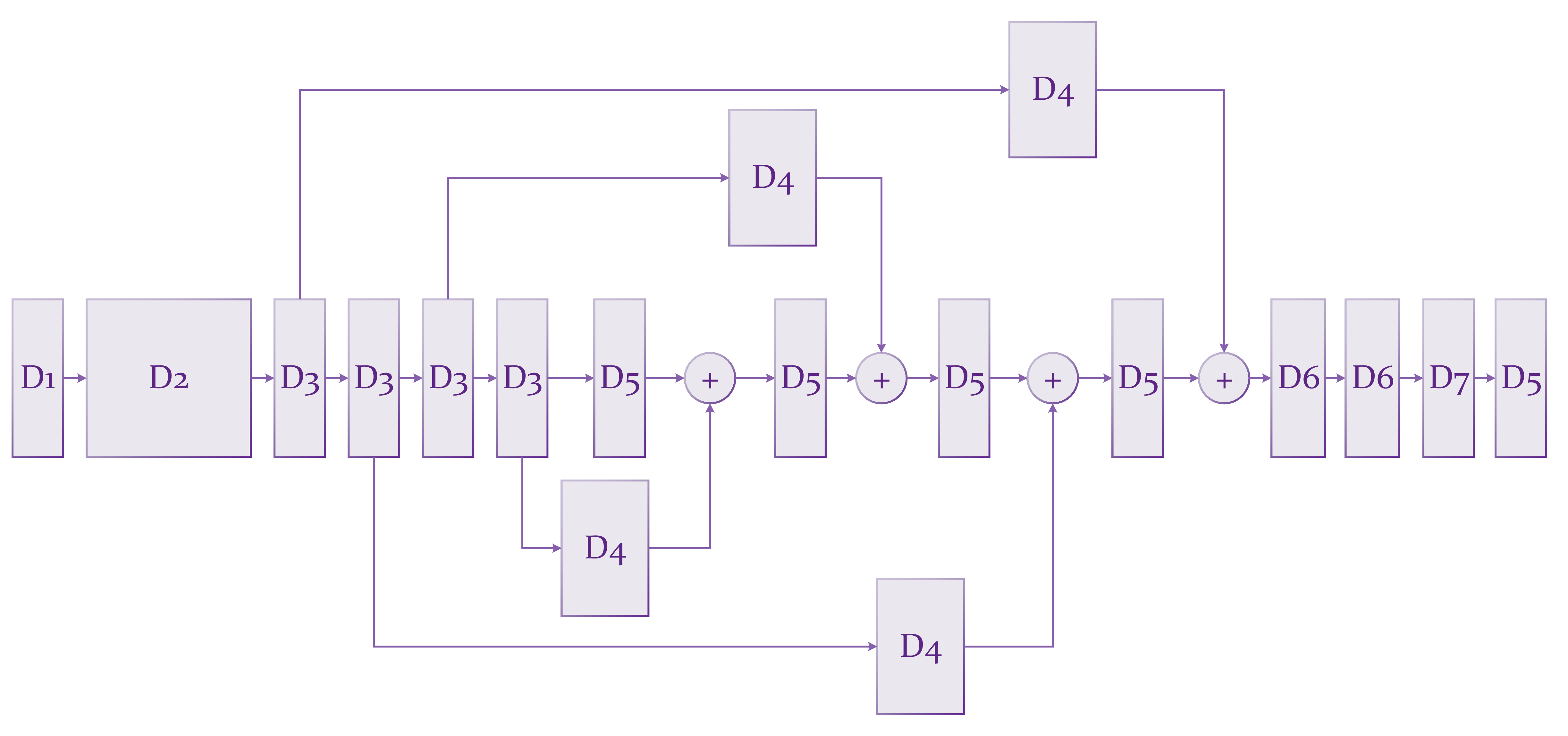}
\caption{The ``hourglass network'' \cite{newell2016stacked} used as the residual regression network. The Blocks D1-D7 are defined in Table \ref{table:RegressionContent}. See also text.}
\label{fig:RegressionContent}
\end{figure}

\textbf{Regression subnetwork.} For the residual regression subnetwork, we used a (slightly) modified ``hourglass network'' \cite{newell2016stacked}, which is a recently proposed state-of-the-art architecture for bottom-up, top-down inference. The network is shown in Fig. \ref{fig:RegressionContent} and Table \ref{table:RegressionContent}. Briefly, the network builds on top of the concepts described in \cite{long2015fully}, improving a few fundamental aspects. The first one is that extends \cite{long2015fully} within residual learning. 
The second one is that instead of passing the lower level futures through a convolution layer with the same number of channels as the final scoring layer, the network passes the features through a set of 3 convolutional blocks that allow the network to reanalyse and learn how to combine features extracted at different resolutions. See \cite{newell2016stacked} for more details. Our modification was in the introduction of deconvolution layers D5 for recovering the lost spatial resolution (as opposed to nearest neighbour upsampling used in \cite{newell2016stacked}). Also, as in the detection network, the output is brought back to the input's resolution using another trained deconvolutional layer D5.

\begin{table}
	\begin{center}
		\caption{Block specification for the ``hourglass network''. Torch notations (channels, kernel, stride) and (kernel, stride) are used to define the conv and pooling layers. The bottleneck modules are defined as in \cite{he2016identity}. }
        \scriptsize
		\label{table:RegressionContent}
		\begin{tabular}{| *7{>{\raggedright\arraybackslash}p{1.7cm}|}}
        \hline
			D1 & D2 & D3 & D4 & D5 & D6  & D7  \\ \hline
		 1x conv layer(64, 7x7, 2x2), 1x pooling(2x2,2x2)   & 3x bottleneck modules   & 1x maxpooling (2x2, 2x2), 3x bottleneck modules  & 3x bottleneck modules & 1x deconv. layer (256, 2x2, 2x2)  &  1x conv layer (512, 1x1, 1x1) & 1x conv scoring layer (16, 1x1, 1x1) \\ \hline
		\end{tabular}
	\end{center}
\end{table}
\setlength{\tabcolsep}{1.4pt}

\textbf{Training.} \label{sec:training}
For training on MPII, we applied similar augmentations as before, with the difference being that augmentations were applied randomly. Also, due to memory issues, the input image was rescaled to 256x256px. Again, we started by first training the body part detection network, fine-tuning from ResNet-152 \cite{he2016deep} pre-trained on ImageNet \cite{deng2009imagenet}. The detectors were then trained for about 50 epochs using a learning rate progressively decreasing from $ 1e-3 $ to $ 2.5e-5 $. For the regression ``hourglass'' subnetwork, we froze the learning for the detector layers, training only the regression subnetwork. We let this train for 40 epochs using a learning rate of $ 1e-4 $ and then $ 2.5e-5 $. In the end, the networks were trained jointly for 50 more epochs. While we experimented with different initialization strategies, all of them seemed to produce similar results. For the final model, all layers from the regression subnetwork were zero-initialized, except for the deconvolution layers, which were initialized using bilinear upsampling filters, as in \cite{long2015fully}. The network made use of batch normalization, and  was trained with a batch size of 8. For LSP, we follow the same procedure as the one for VGG-FCN, changing only the number of epochs to 30. The network was implemented and trained using Torch7 \cite{collobert2011torch7}. The code, along with the pretrained models will be published on our webpage.

\section{Results}

\subsection{Overview}

We report results for two sets of experiments on the two most challenging data sets for human pose estimation, namely LSP \cite{johnson2010clustered} and MPII \cite{andriluka20142d}. A summary of our results is as follows:

\begin{itemize}
\item
    We show the benefit of the proposed detection-followed-by-regression cascade over a two-step regression approach, similar to the idea described in \cite{pfister2015flowing}, when implemented with both VGG-FCN and residual architectures.
	\item 
    We provide an analysis of the different components of our network illustrating their importance on overall performance. We show that stacking the part heatmaps as proposed in our work is necessary for achieving high performance, and that this performance is significantly better than that of the part detection network alone.	  
      \item
    We show the benefit of using a residual architecture over VGG-FCN.
    \item   
	We compare the performance of our method with that of recently published methods illustrating that both versions of our cascade achieve top performance on both the MPII and LSP data sets.  
\end{itemize}
  
\subsection{Analysis}
\label{S:analysis}

We carried out a series of experiments in order to investigate the impact of the various components of our architecture on performance. In all cases, training and testing was done on MPII training and validation set, respectively. The results are summarized in Table \ref{table:OursCompMPIIDet}. In particular, we report the performance of 
\begin{enumerate}[i]
\item the overall part heatmap regression (which is equivalent to ``Detection+regression'') for both residual and VGG-FCN architectures.
\item
the residual part detection network alone (Detection only).
\item
the residual detection network but trained to perform direct regression (Regression only).
\item
a two-step regression approach as in \cite{pfister2015flowing} (Regression+regression), but implemented with both residual and VGG-FCN architectures.
\end{enumerate}

We first observe that there is a large performance gap between residual part heatmap regression and the same cascade but implemented with a VGG-FCN. Residual detection alone works well, but the regression subnetwork provides a large boost in performance, showing that using the stacked part heatmaps as input to residual regression is necessary for achieving high performance. 

Furthermore, we observe that direct regression alone (case iii above) performs better than detection alone, but overall our detection-followed-by-regression cascade significantly outperforms the two-step regression approach. Notably, we found that the proposed part heatmap regression is also considerably easier to train. Not surprisingly, the gap between detection-followed-by-regression and two-step regression when both are implemented with VGG-FCN is much bigger. Overall, these results clearly verify the importance of using (a) part detection heatmaps to guide the regression subnetwork and (b) a residual architecture.

\setlength{\tabcolsep}{2pt}
\begin{table}
	\begin{center}
		\caption{Comparison between different variants of the proposed residual architecture on MPII validation set, using PCKh metric. The overall residual part heatmap regression architecture is equivalent to ``Detection+regression''.}
		\label{table:OursCompMPIIDet}
		\begin{tabular}{lllllllll}
			&Head & Shoulder & Elbow & Wrist & Hip & Knee  & Ankle & Total \\
            \hline
		Part heatmap regression(Res)& \textbf{97.3}  & \textbf{95.2}  & \textbf{89.9}  & \textbf{85.3}  & \textbf{89.4}  & \textbf{85.7} & \textbf{81.9} & \textbf{88.2} \\	
            Part heatmap regression(VGG) & 95.6  & 92.2  & 83.5  & 78.3  & 84.5  & 77.3 & 70.0 & 83.2 \\ \hline        	           
			Detection only(Res) & 96.2  & 91.3  & 83.4  & 74.5  & 83.1  & 76.6 & 71.3 & 82.6\\
                        Regression only(Res) & 96.4  & 92.8  & 84.5  & 77.3  & 84.5  & 79.9 & 74.0 & 84.2 \\
            Regression+regression(Res) & 96.7  & 93.6  & 86.1  & 80.1  & 88.1  & 80.5 & 76.7 & 85.7 \\ 
            Regression+regression(VGG) & 92.8  & 85.6  & 77.5  & 70.4  & 73.5  & 69.3 & 66.5 & 76.7 \\
            \hline
		\end{tabular}
	\end{center}
\end{table}
\setlength{\tabcolsep}{1.4pt}

\setlength{\tabcolsep}{2pt}
\begin{table}
	\begin{center}
		\caption{PCKh-based comparison with state-of-the-art on MPII}
        \label{table:MPII}
		\begin{tabular}{lllllllll}
		 &Head & Shoulder & Elbow & Wrist & Hip & Knee  & Ankle & Total \\
		 \hline
		 Part heatmap regression(Res)& \textbf{97.9}  & 95.1  & 89.9  & \textbf{85.3}  & \textbf{89.4}  & \textbf{85.7} & \textbf{81.9} & \textbf{89.7}\\
         Part heatmap regression(VGG)& 96.8  & 91.3  & 82.9  & 77.5  & 83.2  & 74.4 & 67.5 & 82.7\\
         \hline
        Newell et al., arXiv'16 \cite{newell2016stacked} & 97.6 & \textbf{95.4} & \textbf{90.0} & 85.2 & 88.7 & 85.0 & 80.6 & 89.4 \\
        Wei et al., CVPR'16 \cite{wei2016convolutional} & 97.8 & 95.0 & 88.7 & 84.0 & 88.4 & 82.8 & 79.4 & 88.5 \\
        Insafutdinov et al., arXiv'16 \cite{insafutdinov2016deepercut} & 96.6 & 94.6 & 88.5 & 84.4 & 87.6 & 83.9 & 79.4 & 88.3 \\
        Gkioxary et al., arXiv'16 \cite{gkioxari2016chained} & 96.2 & 93.1 & 86.7 & 82.1 & 85.2 & 81.4 & 74.1 & 86.1 \\
        Lifshitz et al., arXiv'16 \cite{lifshitz2016human} & 97.8 & 93.3 & 85.7 & 80.4 & 85.3 & 76.6 & 70.2 & 85.0 \\
        Pishchulin et. al., CVPR'16 \cite{pishchulin2015deepcut} & 94.1 & 90.2 & 83.4 & 77.3 & 82.6 & 75.7 & 68.6 & 82.4 \\
        Hu\&Ramanan., CVPR'16 \cite{hu2015bottom}  & 95.0 & 91.6 & 83.0 & 76.6 & 81.9 & 74.25 & 69.5 & 82.4 \\
        Carreira et al., CVPR'16 \cite{carreira2015human}& 95.7 & 91.7 & 81.7 & 72.4 & 82.8 & 73.2 & 66.4 & 81.3 \\
		 Tompson et al., NIPS'14 \cite{tompson2014joint} & 95.8  & 90.3  & 80.5  & 74.3  & 77.6  & 69.7 & 62.8 & 79.6\\
		 Tompson et al., CVPR'15 \cite{tompson2015efficient} & 96.1  & 91.9  & 83.9  & 77.8  & 80.9  & 72.3 & 64.8 & 82.0 \\
			\hline
		\end{tabular}
	\end{center}
\end{table}
\setlength{\tabcolsep}{1.4pt}

\subsection{Comparison with state-of-the-art}

\begin{figure}
\centering 
\includegraphics[scale=0.18]{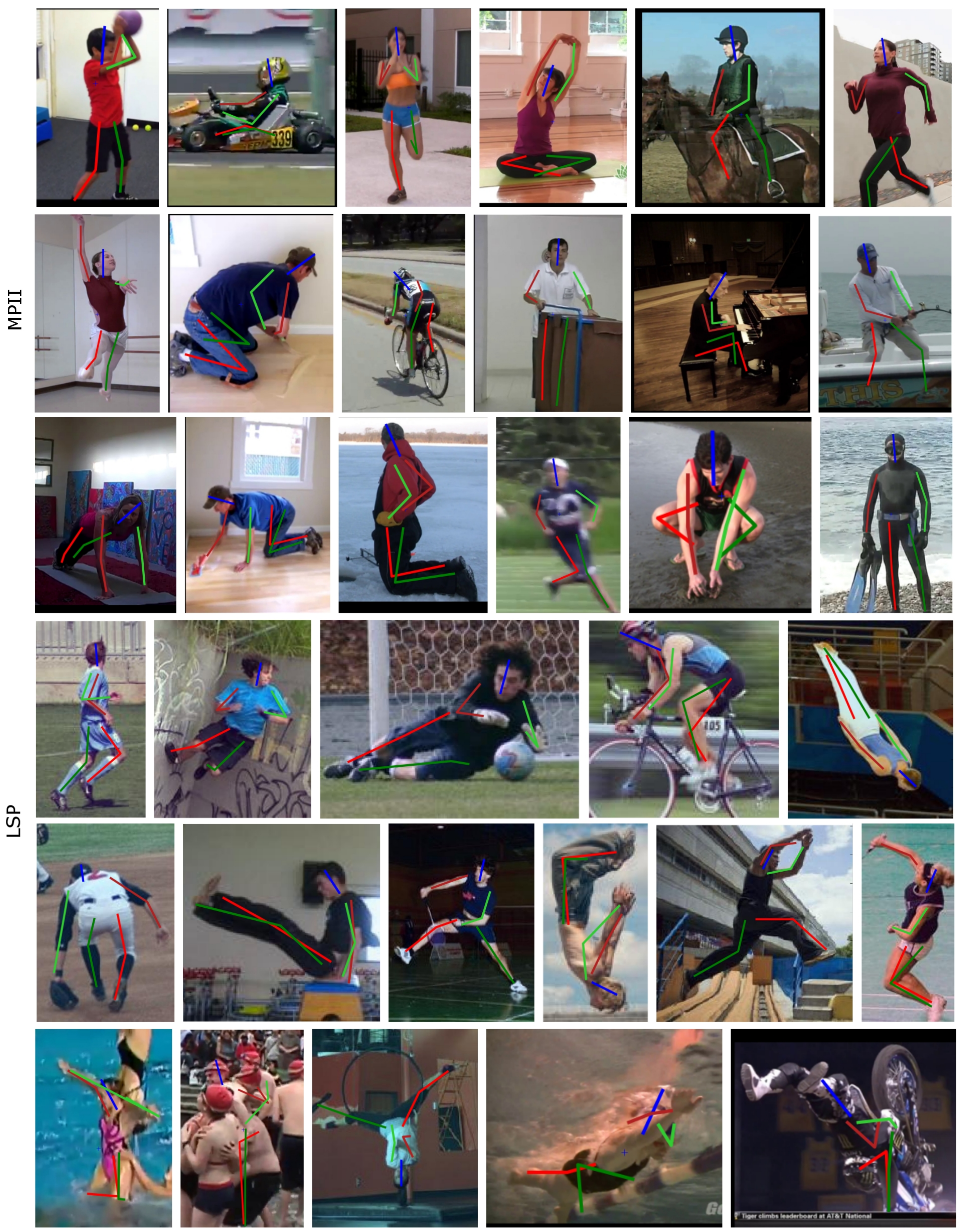}
\caption{Examples of poses obtained using our method on MPII (first 3 rows), and LSP (4th and 5th row). Observe that our method copes well with both occlusions and difficult poses. The last row shows some fail cases caused by combinations of extreme occlusion and rare poses.}
\label{fig:examples}
\end{figure}

In this section, we compare the performance of our method with that of published methods currently representing the state-of-the-art. Tables \ref{table:MPII} and \ref{table:LSP} summarize our results on MPII and LSP, respectively. Our results show that both VGG-based and residual part heatmap regression are very competitive with the latter, along with the other two residual-based architectures \cite{newell2016stacked, insafutdinov2016deepercut}, being top performers on both datasets. Notably, very close in performance is the method of \cite{wei2016convolutional} which is not based on residual learning but performs a sequence of 6 CNN regressions, being also much more challenging to train \cite{wei2016convolutional}. Examples of fitting results from MPII and LSP for the case of residual part heatmap regression can be seen in Fig. \ref{fig:examples}.

\setlength{\tabcolsep}{2pt}
\begin{table}
	\begin{center}
		\caption{PCK-based comparison with the state-of-the-art on LSP}
        \label{table:LSP}
		\begin{tabular}{lllllllll}
		 &Head & Shoulder & Elbow & Wrist & Hip & Knee  & Ankle & Total \\
		 \hline
		 Part heatmap regression(Res) & 96.3  & 92.2  & \textbf{88.2}  & \textbf{85.2}  & \textbf{92.2}  & \textbf{91.5} & 88.6 & \textbf{90.7} \\
         Part heatmap regression(VGG) & 94.8  & 86.6  & 79.5  & 73.5  & 88.1  & 83.2 & 78.5 & 83.5 \\
		 \hline
         Wei et al., CVPR'16 \cite{wei2016convolutional}& \textbf{97.8} & 92.5 & 87.0 & 83.9 & 91.5 & 90.8 & \textbf{89.9} & 90.5 \\
         Insafutdinov et al., arXiv'16 \cite{insafutdinov2016deepercut}& 97.4 & \textbf{92.7} & 87.5 & 84.4 & 91.5 & 89.9 & 87.2& 90.1 \\
         Pishchulin et al.CVPR'16 \cite{pishchulin2015deepcut} & 97.0 & 91.0 & 83.8 & 78.1 & 91.0 & 86.7 & 82.0 & 87.1 \\
         Lifshitz et al., arXiv'16 \cite{lifshitz2016human} & 96.8 & 89.0 & 82.7 & 79.1 & 90.9 & 86.0 & 82.5 & 86.7 \\
         Yang et al., CVPR'16 \cite{yang2016end}& 90.6 &78.1 & 73.8 & 68.8 & 74.8 &69.9 &58.9 &73.6 \\
         Carreira et al., CVPR'16 \cite{carreira2015human}& 90.5 & 81.8 & 65.8 & 59.8 & 81.6 & 70.6 & 62.0 & 73.1 \\
		 Tompson et al., NIPS'14 \cite{tompson2014joint}& 90.6  & 79.2  & 67.9  & 63.4  & 69.5  & 71.0 & 64.2 & 72.3 \\
		 Fan et al., CVPR'15 \cite{Fan_2015_CVPR}& 92.4  & 75.2  & 65.3  & 64.0  & 75.7  & 68.3 & 70.4 & 73.0 \\
		 Chen\&Yuille, NIPS'14 \cite{chen2014articulated}& 91.8  & 78.2  & 71.8  & 65.5  & 73.3  & 70.2 & 63.4 & 73.4 \\
		 
			\hline
		\end{tabular}
	\end{center}
\end{table}
\setlength{\tabcolsep}{1.4pt}

\section{Acknowledgement}

We would like to thank Leonid Pishchulin for graciously producing our results on MPII with unprecedented quickness.

\section{Conclusions}

We proposed a CNN cascaded architecture for human pose estimation particularly suitable for learning part relationships and spatial context, and robustly inferring pose even for the case of severe part occlusions. Key feature of our network is the joint regression of part detection heatmaps. The proposed architecture is very simple and can be trained end-to-end, achieving top performance on the MPII and LSP data sets. 

\clearpage

\bibliographystyle{splncs}
\bibliography{egbib}
\end{document}